\documentclass{llncs}

\usepackage[edges]{forest}

\usepackage{graphicx}
\usepackage{xcolor}
\usepackage{subfigure}
\usepackage{algorithm}
\usepackage{algorithmic}
\usepackage{comment}
\usepackage{scalerel}
\usepackage{float}
\usepackage{booktabs}

\usepackage{array}
\newcolumntype{x}[1]{>{\centering\arraybackslash\hspace{0pt}}p{#1}}

\newcommand{\hendrik}[1]{\textcolor{red}{[HB:#1]}}
\newcommand{\toon}[1]{ }
\newcommand{\ttoon}[1]{\textcolor{blue}{[TVC:#1]}}

\makeatletter
\newcommand{\ALOOP}[1]{\ALC@it\algorithmicloop\ #1%
  \begin{ALC@loop}}
\newcommand{\ENDALOOP}{\end{ALC@loop}\ALC@it\algorithmicendloop}

\makeatother

\begin{document}

\title{COBRAS\textsuperscript{TS}: A new approach to Semi-Supervised Clustering of Time Series}
\titlerunning{COBRAS\textsuperscript{TS}: A new approach to Semi-Supervised Clustering of Time Series}  
\author{Toon Van Craenendonck, Wannes Meert, Sebastijan Duman\v{c}i\'{c} and \\ Hendrik Blockeel}
\authorrunning{Author} 
%
\tocauthor{Authors}
\institute{KU Leuven, Department of Computer Science\\ \{firstname.lastname\}@kuleuven.be }

\maketitle              

\begin{abstract}
Clustering is ubiquitous in data analysis, including analysis of time series. It is inherently subjective: different users may prefer different clusterings for a particular dataset.  Semi-supervised clustering addresses this by allowing the user to provide examples of instances that should (not) be in the same cluster. This paper studies semi-supervised clustering in the context of time series. We show that COBRAS, a state-of-the-art semi-supervised clustering method, can be adapted to this setting. We refer to this approach as COBRAS\textsuperscript{TS}. An extensive experimental evaluation supports the following claims: (1) COBRAS\textsuperscript{TS} far outperforms the current state of the art in semi-supervised clustering for time series, and thus presents a new baseline for the field; (2) COBRAS\textsuperscript{TS} can identify clusters with separated components; (3) COBRAS\textsuperscript{TS} can identify clusters that are characterized by small local patterns; (4) a small amount of semi-supervision can greatly improve clustering quality for time series; (5) the choice of the clustering algorithm matters (contrary to earlier claims in the literature).
\end{abstract}

\toon{maybe add explanation of DTW}
\toon{and cDTW}
\toon{and discussion of plugging other methods into COBRAS (k-Shape)}

\section{Introduction}
\toon{ add a sentence about how ubiquitous time series data is, and hence the need to obtain insights from it? }

Clustering is ubiquitous in data analysis.  There is a large diversity in  algorithms, loss functions, similarity measures, etc.  This is partly due to the fact that clustering is inherently subjective: in many cases, there is no single “correct” clustering, and different users may prefer different clusterings, depending on their goals and prior knowledge \cite{Caruana06metaclustering,ScienceOrArt}.  Depending on their preference, they should use the right algorithm, similarity measure, loss function, hyperparameter settings, etc.  This requires a fair amount of knowledge and expertise on the user's side.

Semi-supervised clustering methods deal with this subjectiveness in a different manner. They allow the user to specify constraints that express their subjective interests \cite{Wagstaff01constrainedk-means}. These constraints can then guide the algorithm towards solutions that the user finds interesting. Many such systems obtain these constraints by asking the user to answer queries of the following type: \emph{should these two elements be in the same cluster?} A must-link constraint is obtained if the answer is yes, a cannot-link otherwise. In many situations, answering this type of questions is much easier for the user than selecting the right algorithm, defining the similarity measure, etc.

In the context of clustering {\em time series}, the subjectiveness of clustering is even more prominent. In some contexts, the time scale matters, in other contexts it does not. Similarly, the response scale may (not) matter.  One may want to cluster time series based on certain types of qualitative behavior (monotonic, periodic, \ldots), local patterns that occur in them, etc.  
Despite this variability, and although there is a plethora of work on time series clustering, semi-supervised clustering of time series has only very recently started receiving attention. The cDTW\textsuperscript{SS} method developed by Dau et al.\ \cite{KeoghSS} is to our knowledge the only attempt to date to address this task.


In this paper, we show that COBRAS, an existing semi-supervised clustering system, can be used practically ``as-is'' for time series clustering. The only adaptation that is needed, is the plugging in of a suitable similarity measure and a corresponding (unsupervised) clustering approach for time series.  Two plug-in methods are considered for this: spectral clustering using dynamic time warping (DTW), and k-Shape \cite{kshape}.  We refer to COBRAS with one of these plugged in as COBRAS\textsuperscript{TS} (COBRAS for Time Series).  We perform an extensive experimental evaluation of this approach.

The main contributions of the paper are twofold.
First, it contributes a novel approach to semi-supervised clustering of time series, and two concrete, freely downloadable and ready-to-use implementations of it.
Second, the paper provides extensive evidence for the following claims: (1) COBRAS\textsuperscript{TS} outperforms cDTW\textsuperscript{SS} (the current state of the art) by a large margin; (2) COBRAS\textsuperscript{TS} can identify clusters with separated components, and this is one reason why it performs well; (3) COBRAS\textsuperscript{TS} can identify clusters that are characterized by small local patterns; (4) a small amount of supervision can greatly improve results in time series clustering; (5) the choice of clustering algorithm matters, it is not negligible compared to the choice of similarity.  Except for claim 4, all these claims are novel, and some are at variance with the current literature.  Claim 4 has been made before, but with much weaker empirical support.

\section{Related work}
\label{sec:relatedwork}
Semi-supervised clustering has been studied extensively for clustering attribute-value data, starting with COP-KMeans \cite{Wagstaff01constrainedk-means}. Most semi-supervised methods extend unsupervised ones by adapting their clustering procedure \cite{Wagstaff01constrainedk-means}, their similarity measure \cite{xing2002distance}, or both \cite{basu:sdm04}. Alternatively, constraints can also be used to select and tune an unsupervised clustering algorithm \cite{COBS}. 

Traditional methods assume that a set of pairwise queries is given prior to running the clustering algorithm, and in practice, pairs are often queried randomly. Active semi-supervised clustering methods try to query the most informative pairs first, instead of random ones \cite{Mallapragada2008}. Typically, this results in better clusterings for an equal number of queries. COBRAS \cite{COBRAS} is a recently proposed active semi-supervised clustering method that was shown to be effective for clustering attribute-value data. 
In this paper, we show that it can be used to cluster time series with little modification. We describe COBRAS in more detail in the next section. 

In contrast to the wealth of papers on semi-supervised clustering of attribute-value data, only one method has been proposed specifically for semi-supervised time series clustering. cDTW\textsuperscript{SS} \cite{KeoghSS} uses pairwise constraints to tune the warping width parameter $w$ in constrained DTW. cDTW\textsuperscript{SS} is also an active clustering method, as it comes with a strategy to select the pairwise queries that are most informative for tuning $w$. We compare COBRAS\textsuperscript{TS} to this method in the experiments.

Zhou et al.\ \cite{Zhou2015} introduce a method that uses different distance measures to generate pairwise constraints, and then uses these constraints in a semi-supervised variant of spectral clustering \cite{Kulis2009}. While related, this is not a semi-supervised method, as it does not exploit supervision by the user. Rather, it makes it possible to use semi-supervised algorithms in an unsupervised setting.

In contrast to semi-supervised time series clustering, semi-supervised time series \emph{classification} has received significant attention \cite{Wei2006KDD}.  Note that these two settings are quite different: in semi-supervised classification, the set of classes is known beforehand, and at least one labeled example of each class is provided. In semi-supervised clustering, it is not known in advance how many classes (clusters) there are, and a class may be identified correctly even if none of its instances have been involved in the pairwise constraints.  

\section{Clustering time series with COBRAS}
\label{sec:algo}


\subsection{COBRAS}
\label{sec:cobras_hl}

We describe COBRAS only to the extent necessary to follow the remainder of the paper; for more information, see Van Craenendonck et al. \cite{COBRA,COBRAS}.

COBRAS is based on two key ideas. The first \cite{COBRA} is that of \emph{super-instances}: sets of instances that are temporarily assumed to belong to the same cluster in the unknown target clustering. In COBRAS, a clustering is a set of clusters, each cluster is a set of super-instances, and each super-instance is a set of instances. This intermediate level of super-instances makes it possible to exploit constraints much more efficiently: querying is performed at the level of super-instances, which means that each instance does not have to be considered individually in the querying process. The second key idea in COBRAS \cite{COBRAS} is that of the \emph{automatic detection of the right level} at which these super-instances are constructed.  For this, it uses an iterative refinement process.  COBRAS starts with a single super-instance that contains all the examples, and a single cluster containing that super-instance.  In each iteration the largest super-instance is taken out of its cluster, split into smaller super-instances, and the latter are reassigned to (new or existing) clusters. Thus, COBRAS constructs a clustering of super-instances at an increasingly fine-grained level of granularity. The clustering process stops when the query budget is exhausted.

\begin{figure}[t!]
\centering     
\includegraphics[width=1.0\textwidth]{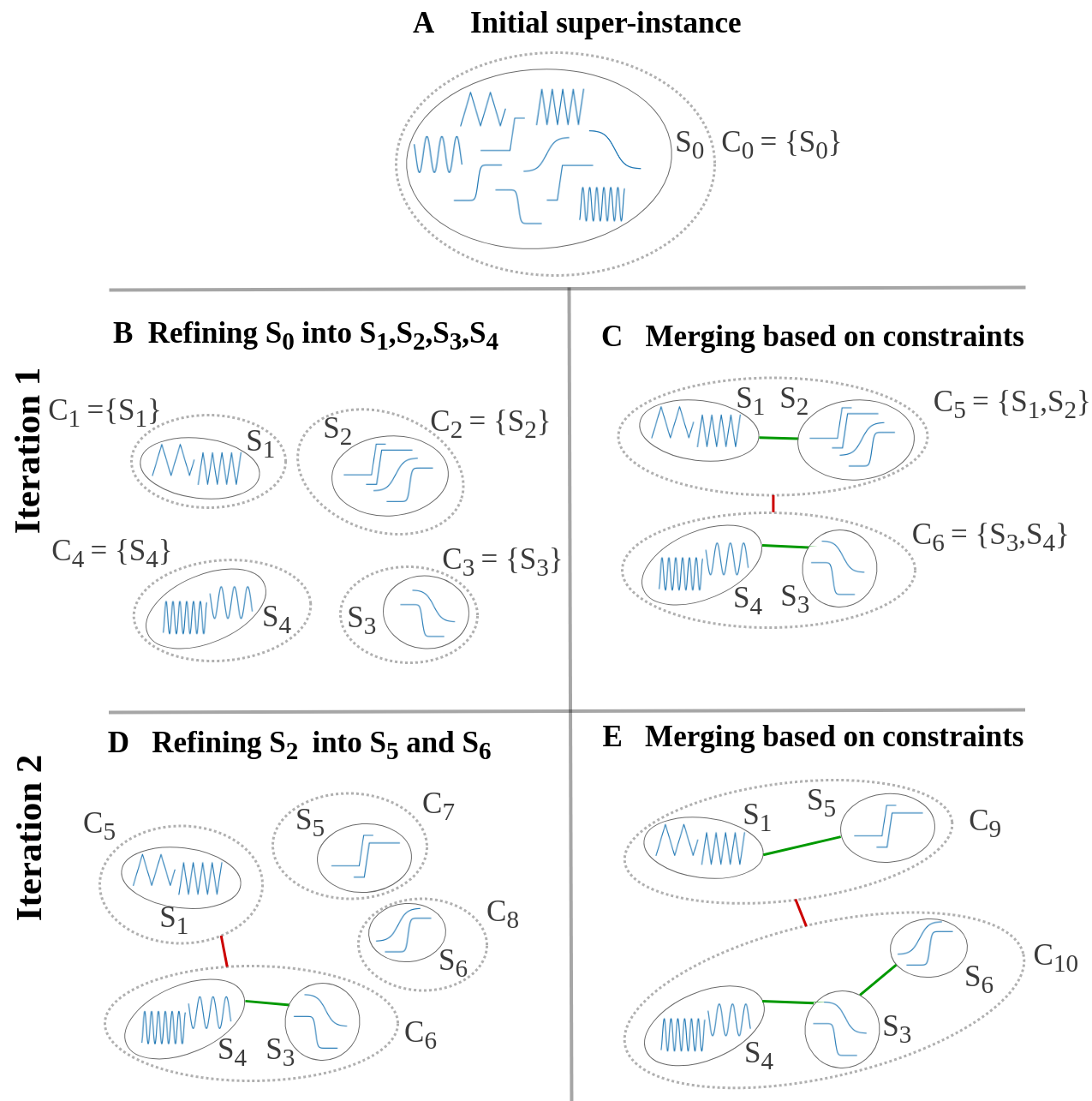}  
\caption{An illustration of the COBRAS clustering procedure.}\label{fig:cobras_proc}
\end{figure}

We illustrate this procedure using the example in Figure \ref{fig:cobras_proc}. Panel A shows a toy dataset that can be clustered according to several criteria.  We consider differentiability and monotonicity as relevant properties. Initially, all instances belong to a single super-instance ($S_0$), which constitutes the only cluster ($C_0$). The second and third rows of Figure \ref{fig:cobras_proc} show two iterations of COBRAS. 

In the first step of iteration 1, COBRAS refines $S_0$ into 4 new super-instances, which are each put in their own cluster (panel B). The refinement procedure uses $k$-means, and the number of super-instances in which to split is determined based on constraints; for details, see \cite{COBRAS}. In the second step of iteration 1, COBRAS determines the relation between new and existing clusters. To determine the relation between two clusters, COBRAS queries the pairwise relation between the medoids of their closest super-instances. In this example, we assume that the user is interested in a clustering based on differentiability. The relation between $C_1 = \{S_1\}$ and $C_2 = \{S_2\}$ is determined by posing the following query to the user: \emph{should \raisebox{-.5ex}{\includegraphics[height=2.5ex]{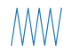}} and \raisebox{-.5ex}{\includegraphics[height=2.5ex]{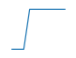}} be in the same cluster?}
The user answers with a must-link constraint, resulting in $C_1$ and $C_2$ being merged into $C_5$. Similarly, COBRAS determines the other pairwise relations between clusters. It does not need to query all of them, as many can be derived through transitivity or entailment \cite{COBRAS}. The first iteration ends once all pairwise relations between clusters are known. This is the situation depicted in panel C. Note that COBRAS has not produced a perfect clustering at this point, as $S_2$ contains both differentiable and non-differentiable instances.

In the second iteration, COBRAS again starts by refining its largest super-instance. In this case, $S_2$ is refined into $S_5$ and $S_6$, as illustrated in panel D. A new cluster is created for each of these super-instances, and the relation between new and existing clusters is determined by querying pairwise constraints. A must-link constraint between $S_5$ and $S_1$ results in the creation of $C_9 = \{S_1,S_5\}$. Similarly, a must-link between $S_6$ and $S_3$ results in the creation of $C_{10} = \{S_3,S_4,S_6\}$. At this point, the second iteration ends as all pairwise relations between clusters are known.

In general, COBRAS keeps repeating its two steps (refining super-instances and querying their pairwise relations) until the query budget is exhausted. 

\subsubsection{Separated components}
\label{sec:sepcomp}
\begin{figure}[t]
  \centering
  \includegraphics[]{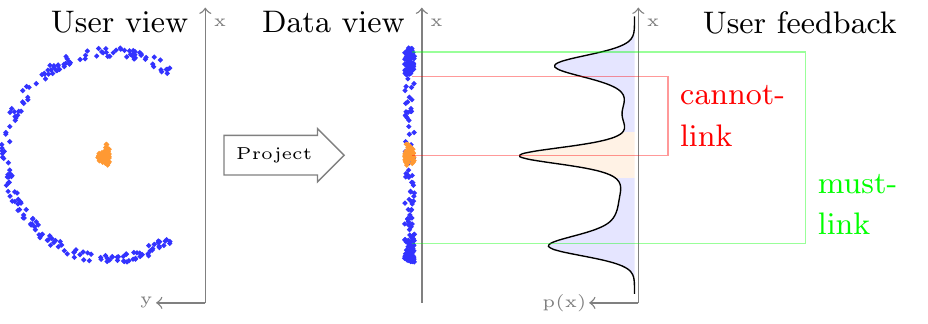}
  \caption{Clusters may contain separated components when projected on a lower-dimensional subspace.}
  \label{fig:wannes}
\end{figure}

A noteworthy property of COBRAS is that, by interleaving splitting and merging, it can split off a subcluster from a cluster and reassign it to another cluster.  In this way, it can construct clusters that contain {\em separated components} (different dense regions that are separated by a dense region belonging to another cluster).  It may, at first, seem strange to call such a structure a ``cluster'', as clusters are usually considered to be coherent high-density areas.  However, note that a coherent cluster may become incoherent when projected onto a subspace.  Figure~\ref{fig:wannes} illustrates this.  Two clusters are clearly visible in the XY-space, yet projection on the X-axis yields a trimodal distribution where the outer modes belong to one cluster and the middle mode to another.  In semi-supervised clustering, it is realistic that the user evaluates similarity on the basis of more complete information than explicitly present in the data; coherence in the user's mind may therefore not translate to coherence in the data space. 

The need for handling clusters with multi-modal distributions has been mentioned repeatedly in work on time series anomaly detection \cite{6766252}, on unsupervised time series clustering \cite{k-multishape}, and on attribute-value semi-supervised constrained clustering \cite{smieja2017constrained}.  Note, however, a subtle difference between having a multi-modal distribution and containing separated components: the first assumes that the components are separated by a low-density area, whereas the second allows them to be separated by a dense region of instances from another cluster.  




\subsection{COBRAS\textsuperscript{DTW} and COBRAS\textsuperscript{k-Shape}}
\label{sec:cobras_limited}
COBRAS is not suited out-of-the-box for time series clustering, for two reasons. First, it defines the super-instance medoids w.r.t.\ the \emph{Euclidean} distance, which is well-known to be suboptimal for time series. Second, it uses k-means to refine super-instances, which is known to be sub-state-of-the-art for time series clustering \cite{kshape}. 

Both of these issues can easily be resolved by plugging in distance measures and clustering methods that are developed specifically for time series. We refer to this approach as COBRAS\textsuperscript{TS}. We now present two concrete instantiations of it: COBRAS\textsuperscript{DTW} and COBRAS\textsuperscript{k-Shape}.

\subsubsection*{COBRAS\textsuperscript{DTW}}
uses DTW as its distance measure, and spectral clustering to refine super-instances. It is described in Algorithm \ref{algo:COBRAS}. DTW is commonly accepted to be a competitive distance measure for time series analysis \cite{Bagnall2017}, and spectral clustering is well-known to be an effective clustering method \cite{vonLuxburg2007}. We use the constrained variant of DTW, cDTW, which restricts the amount by which the warping path can deviate from the diagonal in the warping matrix. cDTW offers benefits over DTW in terms of both runtime and solution quality \cite{kshape,KeoghSS}, if run with an appropriate window width.

\begin{algorithm}[ht]
\caption{COBRAS\textsuperscript{DTW}}
\label{algo:COBRAS}
\begin{algorithmic}[1]
 \REQUIRE A dataset, the DTW warping window width $w$, the $\gamma$ parameter used in converting distances to similarities and
 access to an oracle answering pairwise queries
  \ENSURE A clustering
\STATE Compute the full pairwise DTW distance matrix
\STATE Convert each distance $d$ to an affinity $a$: $a_{i,j} = e^{-\gamma d_{i,j}}$
\STATE Run COBRAS, substituting k-means for splitting super-instances with spectral clustering on the previously computed affinity matrix
\end{algorithmic}    
\end{algorithm}

\subsubsection*{COBRAS\textsuperscript{k-Shape}} 
uses the shape-based distance (SBD, \cite{kshape}) as its distance measure, and the corresponding k-Shape clustering algorithm \cite{kshape} to refine super-instances. k-Shape can be seen as a k-means variant developed specifically for time series. It uses SBD instead of the Euclidean distance, and comes with a method of computing cluster centroids that is tailored to time series. k-Shape was shown to be an effective and scalable method for time series clustering in \cite{kshape}. Instead of the medoid, COBRAS\textsuperscript{k-Shape} uses the instance that is closest to the SBD centroid as a super-instance representative.

\section{Experiments}
\label{sec:experiments}

In our experiments we evaluate COBRAS\textsuperscript{DTW} and COBRAS\textsuperscript{k-Shape} in terms of both clustering quality and runtime, and compare them to state-of-the-art semi-supervised (cDTW\textsuperscript{SS} and COBS) and unsupervised (k-Shape and k-MS) competitors. The experiments presented in this paper are fully reproducible: we provide code for COBRAS\textsuperscript{TS} in a public git repository\footnote{\url{https://bitbucket.org/toon\_vc/cobras\_ts} or using \texttt{pip install cobras\_ts}}, and a separate git repository that contains our scripts for running the experiments\footnote{\url{https://bitbucket.org/toon\_vc/cobras\_ts\_experiments}}. The experiments are performed on the public UCR time series collection \cite{UCRArchive}.

\subsection{Methods}

\subsubsection*{COBRAS\textsuperscript{TS}}
\begin{figure*}
\centering     
\subfigure{\label{fig:gamma_sensitivity}\includegraphics[width=0.49\textwidth]{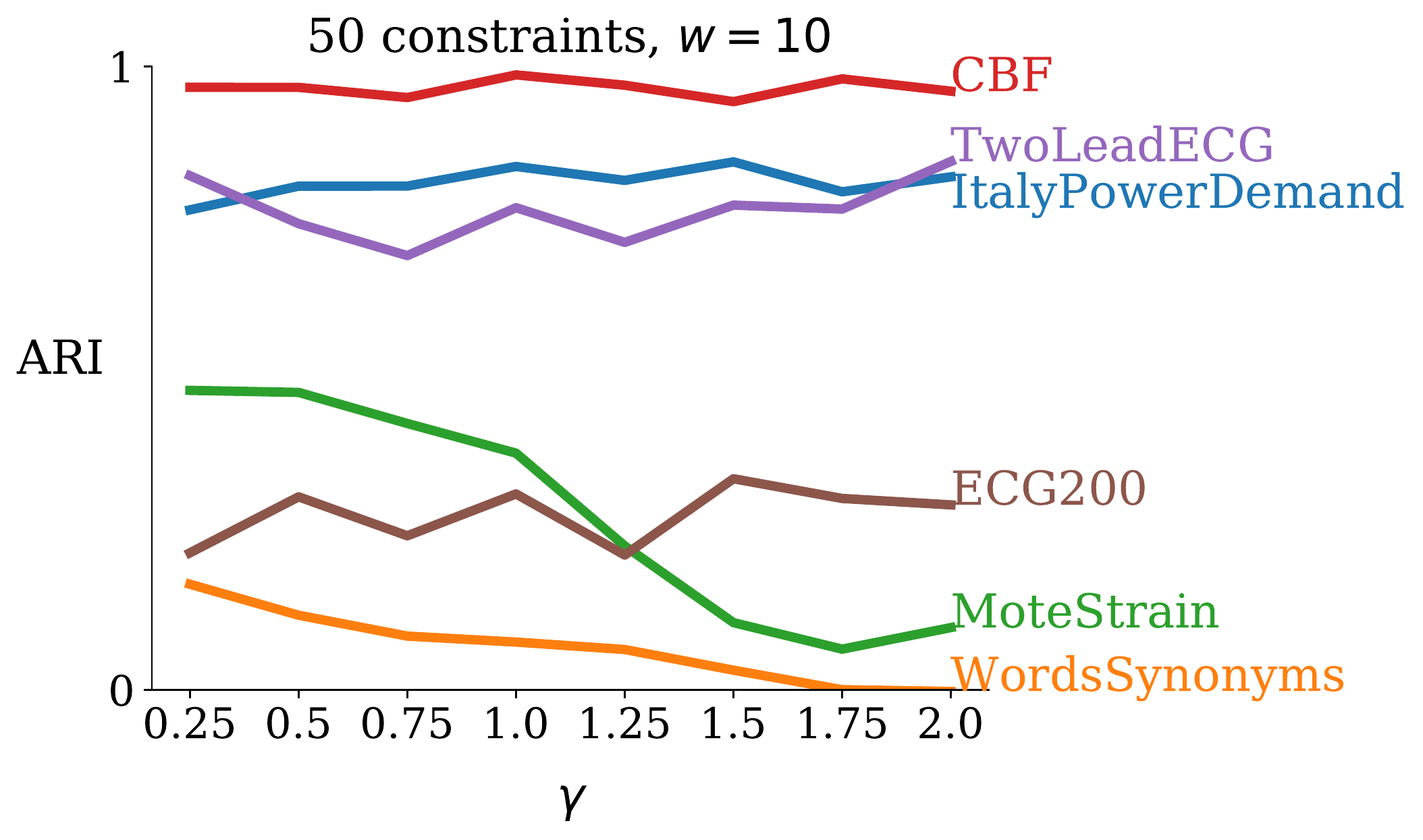}}
\subfigure{\label{fig:w_sensitivity}\includegraphics[width=0.49\textwidth]{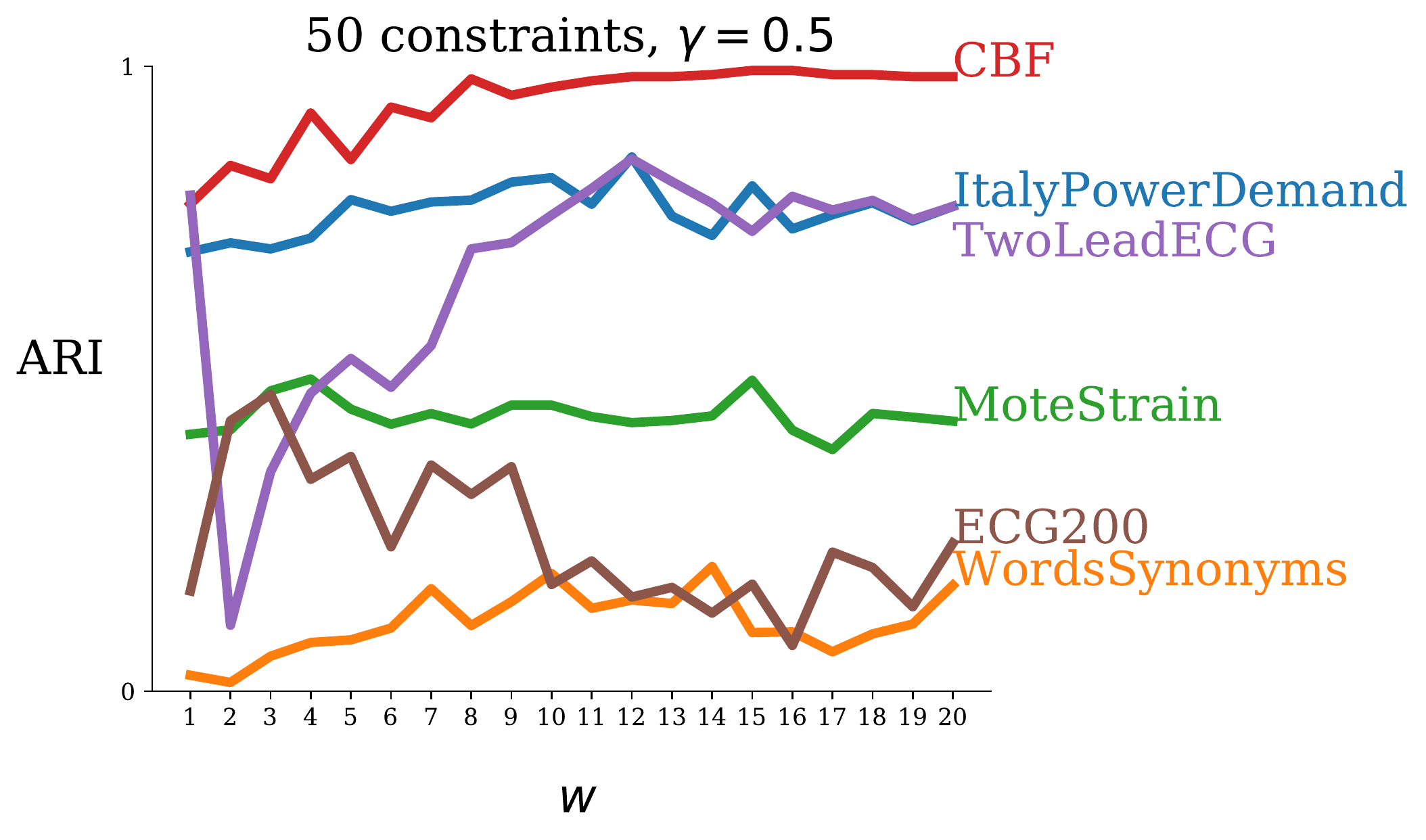}}
\caption{Sensitivity to $\gamma$ and $w$ for several datasets.}\label{fig:param_sens}
\end{figure*}

COBRAS\textsuperscript{k-Shape} has no parameters (the number of clusters used in k-Shape to refine super-instances is chosen based on the constraints in COBRAS). We use a publicly available Python implementation\footnote{\url{https://github.com/Mic92/kshape}} to obtain the k-Shape clusterings. COBRAS\textsuperscript{DTW} has two parameters: $\gamma$ (used in converting distances to affinities) and $w$ (the warping window width). We use a publicly available C implementation to construct the DTW distance matrices \cite{wannes_meert_2018_1202379}. In our experiments, $\gamma$ is set to 0.5 and $w$ to 10\% of the time series length. The value  $w=10\%$ was chosen as Dau et al.\ \cite{KeoghSS} report that most datasets do not require $w$ greater than 10\%. We note that $\gamma$ and $w$ could in principle also be tuned for COBRAS\textsuperscript{DTW}. There is, however, no well-defined way of doing this. We cannot use the constraints for this, as they are actively selected during the execution of the algorithm (which of course requires the affinity matrix to already be constructed). We did not do any tuning on these parameters, as this is also hard in a practical clustering scenario, but observed that the chosen parameter values already performed very well in the experiments. We performed a parameter sensitivity analysis, illustrated in Figure \ref{fig:param_sens}, which shows that the influence of these parameters is highly dataset-dependent: for many datasets their values do not matter much, for some they result in large differences. 

\subsubsection*{cDTW\textsuperscript{SS}}
cDTW\textsuperscript{SS} uses pairwise constraints to tune the $w$ parameter in cDTW. In principle, the resulting tuned cDTW measure can be used with any clustering algorithm. The authors in \cite{KeoghSS} use it in combination with TADPole \cite{Begum:2015:ADT:2783258.2783286}, and we do the same here. We use the code that is publicly available on the authors' website\footnote{\url{https://sites.google.com/site/dtwclustering/}}. The cutoff distances used in TADPole were obtained from the authors in personal communication.

\subsubsection*{COBS}
COBS \cite{COBS} uses constraints to select and tune an unsupervised clustering algorithm. It was originally proposed for attribute-value data, but it can trivially be modified to work with time series data as follows. First, the full pairwise distance matrix is generated with cDTW using $w=10\%$ of the time series length. Next, COBS generates clusterings by varying the hyperparameters of several standard unsupervised clustering methods, and selects the resulting clustering that satisfies the most pairwise queries. We use the active variant of COBS, as described in \cite{COBS}. Note that COBS is conceptually similar to cDTW\textsuperscript{SS}, as both methods use constraints for hyperparameter selection. The important difference is that COBS uses a fixed distance measure and selects and tunes the clustering algorithm, whereas cDTW\textsuperscript{SS} tunes the similarity measure and uses a fixed clustering algorithm. We use the following unsupervised clustering methods and corresponding hyperparameter ranges in COBS: spectral clustering ($K \in [\max(2,K_{true}-5),K_{true}+5]$), hierarchical clustering ($K \in [\max(2,K_{true}-5),K_{true}+5]$, with both average and complete linkage), affinity propagation ($\texttt{damping} \in [0.5,1.0]$) and DBSCAN ($\epsilon \in [\texttt{min pairwise dist.}, \texttt{max. pairwise dist}], \texttt{min\_samples} \in [2,21]$). For the continuous parameters, clusterings were generated for 20 evenly spaced values in the specified intervals. Additionally, the $\gamma$ parameter in converting distances to affinities was varied in $[0,2.0]$ for clustering methods that take affinities as input, which are all of them except DBSCAN, which works with distances. We did not vary the warping window width $w$ for generating clusterings in COBS. This would mean a significant further increase in computation time, both for generating the DTW distance matrices, and for generating clusterings with all methods and parameter settings for each value of $w$.
   
\subsubsection*{k-Shape and k-MS}
Besides the three previous semi-supervised methods, we also include k-Shape \cite{kshape} and k-MultiShape (k-MS) \cite{k-multishape} in our experiments as unsupervised baselines. k-MS \cite{k-multishape} is similar to k-Shape, but uses multiple centroids, instead of one, to represent each cluster. It was found to be the most accurate method in an extensive experimental study that compares a large number of unsupervised time series clustering methods on the UCR collection  \cite{k-multishape}. The number of centroids that k-MS uses to represent a cluster is a parameter; following the original paper we set it to 5 for all datasets. The k-MS code was obtained from the authors. 

\subsection{Data}
We perform experiments on the entire UCR time series classification collection \cite{UCRArchive}, which is the largest public collection of time series datasets. It consists of 85 datasets from a wide variety of domains. The UCR datasets come with a predefined training and test set. We use the test sets as our datasets as they are often much bigger than the training sets. This means that whenever we refer to a dataset in the remainder of this text, we refer to the test set of that dataset as defined in \cite{UCRArchive}. This procedure was also followed by Dau et al. \cite{KeoghSS}.  

 As is typically done in evaluating semi-supervised clustering methods, the classes are assumed to represent the clusterings of interests. When computing rankings and average ARIs, we ignored results from 21 datasets where cDTW\textsuperscript{SS} either crashed or timed out after 24h.\footnote{These datasets are listed at \url{https://bitbucket.org/toon\_vc/cobras\_ts\_experiments}
}

\subsection{Methodology}

We use 10-fold cross-validation, as is common in evaluating semi-supervised clustering methods \cite{Basu2004KDD,Mallapragada2008}. The full dataset is clustered in each run, but the methods can only query pairs of which both instances are in the training set. The result of a run is evaluated by computing the Adjusted Rand Index (ARI) \cite{ARI} on the instances of the test set. The ARI measures the similarity between the generated clusterings and the ground-truth clustering, as indicated by the class labels. It is 0 for a random clustering, and 1 for a perfect one. The final ARI scores that are reported are the average ARIs over the 10 folds. 

We ensure that cDTW\textsuperscript{SS} and COBS do not query pairs that contain instances from the test set by simply excluding such candidates from the list of constraints that they consider. For COBRAS\textsuperscript{TS}, we do this by only using training instances to compute the super-instance representatives. 

COBRAS\textsuperscript{TS} and COBS do not require the number of clusters as an input parameter, whereas cDTW\textsuperscript{SS}, k-Shape and k-MS do. The latter three were given the correct number of clusters, as indicated by the class labels. Note that this is a significant advantage for these algorithms, and that in many practical applications the number of clusters is not known beforehand.

\subsection{Results}

\subsubsection*{Clustering quality}
\begin{figure}[H]
\centering     
\subfigure[]{\label{fig:rank_comparison}\includegraphics[width=0.46\textwidth]{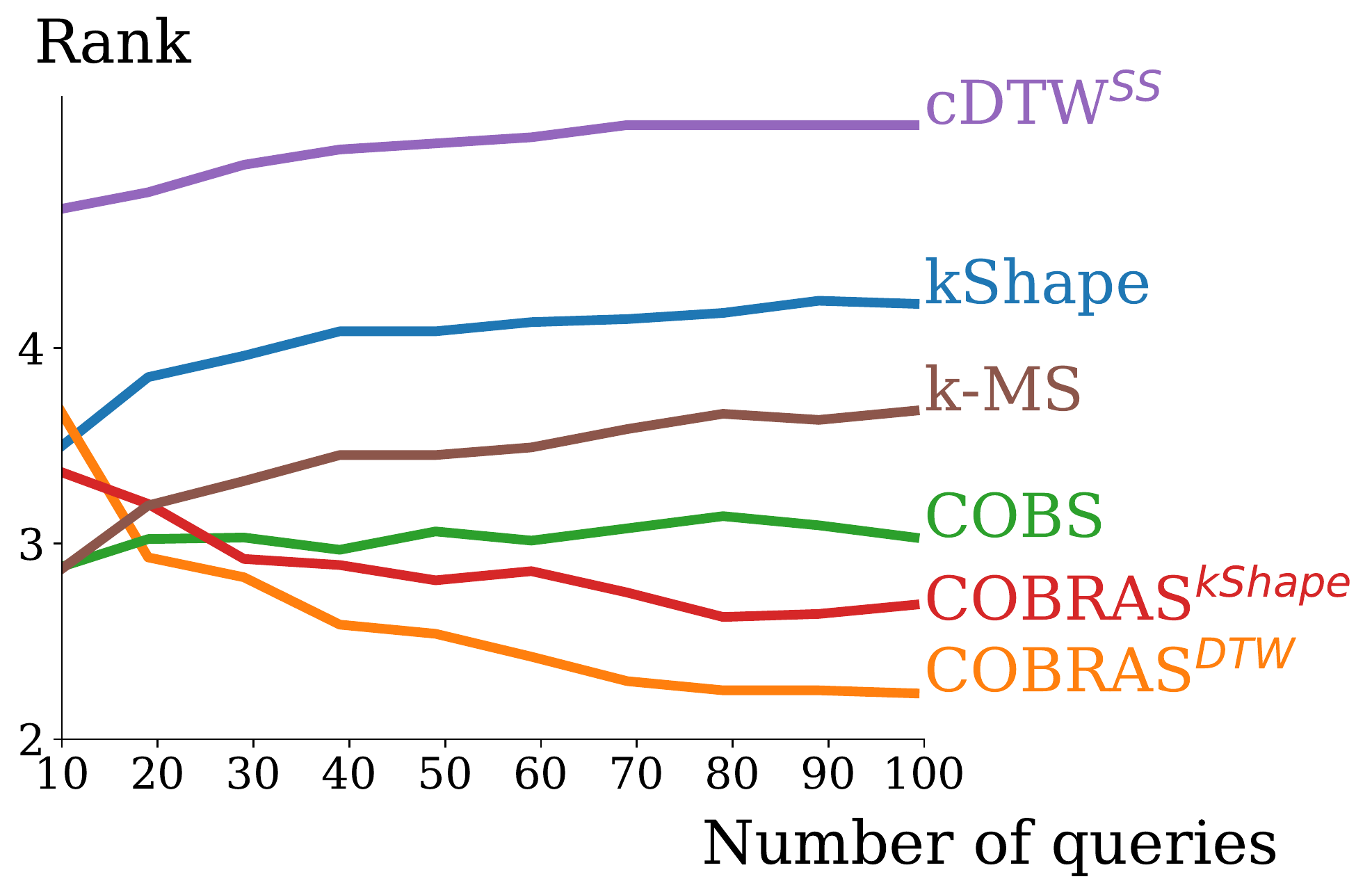}}
\hspace*{15pt}
\subfigure[ ]{\label{fig:ari_comparison}\includegraphics[width=0.46\textwidth]{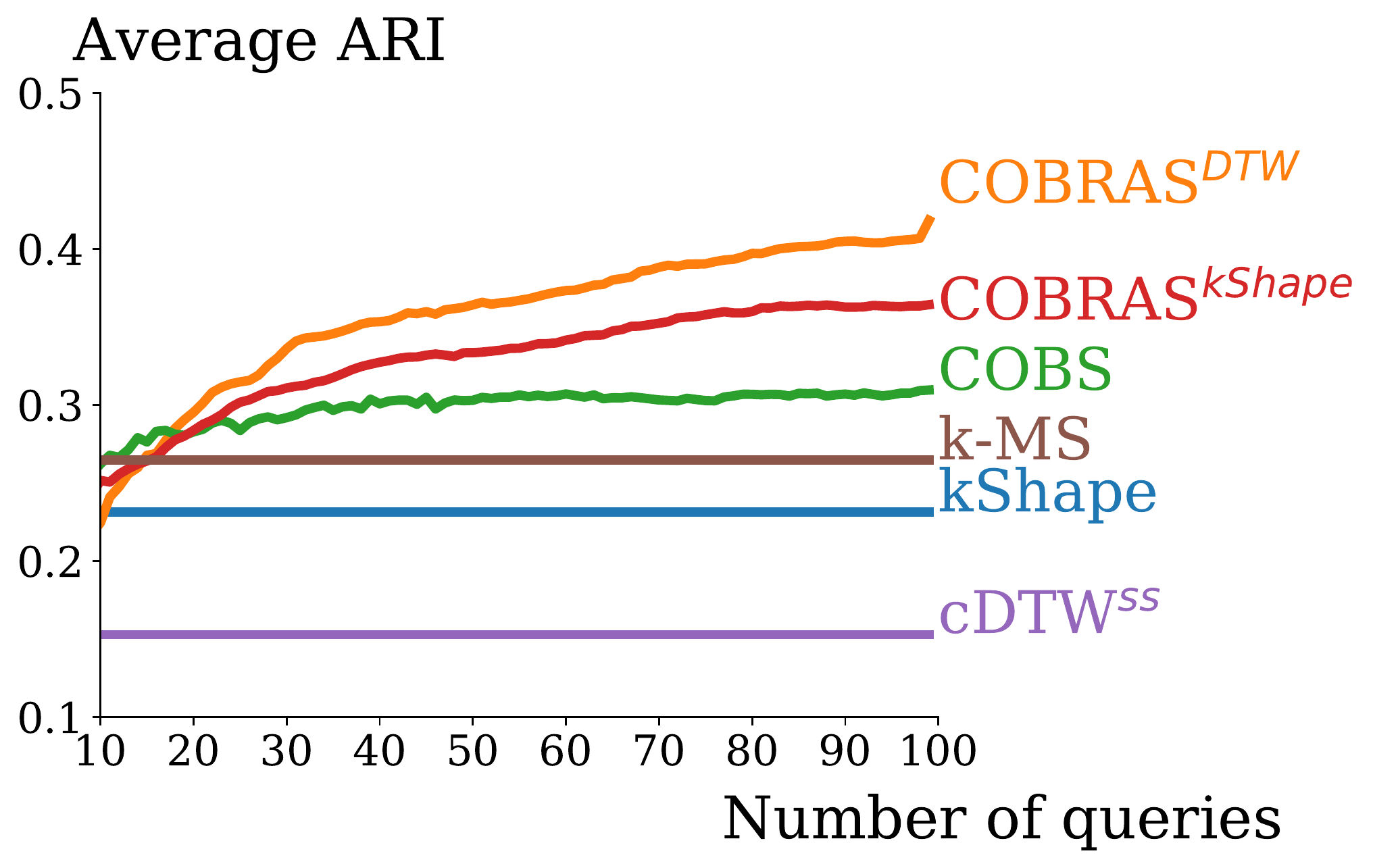}}
\caption{(a) Average rank for all methods over all clustering tasks. Lower is better. (b) Average ARI. Higher is better.}
\end{figure}

Figure \ref{fig:rank_comparison} shows the average ranks of the compared methods over all datasets. Figure \ref{fig:ari_comparison} shows the average ARIs. Both plots clearly show that, on average, COBRAS\textsuperscript{TS} outperforms all the competitors by a large margin. Only when the number of queries is small (roughly $< 15$), is it outperformed by COBS and k-MS. 

These observations are confirmed by Table \ref{table:winloss}, which reports the number of times COBRAS\textsuperscript{DTW} wins and loses against the alternatives. The differences with cDTW\textsuperscript{SS} and k-Shape are significant for all the considered numbers of queries (Wilcoxon test, $p < 0.05$). The difference between COBRAS\textsuperscript{DTW} and COBS is significant for 50 and 100 queries, but not for 25. The same holds for COBRAS\textsuperscript{DTW} vs.\ k-MS.  This confirms the observation from Figure \ref{fig:rank_comparison}, which showed that the performance gap between COBRAS\textsuperscript{DTW} and the competitors becomes larger as more queries are answered. The difference between COBRAS\textsuperscript{DTW} and COBRAS\textsuperscript{k-Shape} is only statistically significant for 100 queries.

\begin{table*}\footnotesize
\centering
  \caption{Wins and losses over the 64 datasets. An asterisk indicates that the difference is significant according to the Wilcoxon test with $p < 0.05$. } \label{table:winloss}
  \begin{tabular}{c  x{1cm}  x{1cm}  x{1cm}  x{1cm}  x{1cm}  x{1cm}  }
  \toprule
\multicolumn{1}{c}{} & \multicolumn{2}{c}{25 queries } & \multicolumn{2}{c}{50 queries}  & \multicolumn{2}{c}{100 queries}  \\ 
\multicolumn{1}{c}{}  & win & loss & win & loss & win & loss \\ 
\midrule
COBRAS\textsuperscript{DTW} vs. COBRAS\textsuperscript{k-Shape} & \textbf{35} & 29 &  \textbf{37} & 27 & \textbf{41*} & 23 \\
COBRAS\textsuperscript{DTW} vs. k-MS  & \textbf{35} & 29 &  \textbf{40*} & 24 & \textbf{47*} & 14 \\
COBRAS\textsuperscript{DTW} vs. COBS  & \textbf{37} & 27 &  \textbf{42*} & 22 &  \textbf{45*} & 19 \\
COBRAS\textsuperscript{DTW} vs. cDTW\textsuperscript{SS}  & \textbf{62*} & 2 & \textbf{53*}   & 11 &  \textbf{55*} & 9  \\
COBRAS\textsuperscript{DTW} vs. k-Shape  & \textbf{40*} & 24 &  \textbf{46*} & 18 & \textbf{50*} & 14 \\
 \bottomrule
  \end{tabular}
\end{table*}

It is surprising to see that the unsupervised baselines significantly outperform the semi-supervised cDTW\textsuperscript{SS}. This conclusion is at variance with the claim that the choice of $w$ dwarfs any improvements by the k-Shape algorithm \cite{KeoghSS}. To ensure that this is not an effect of the evaluation strategy (10-fold CV using the ARI, compared to no CV and the Rand index (RI) in \cite{KeoghSS}), we have also computed the RIs for all of the clusterings generated by k-Shape and compared them directly to the values provided by the authors of cDTW\textsuperscript{SS} on their webpage\footnote{\url{https://sites.google.com/site/dtwclustering/}}. In this experiment k-Shape attained an average RI of 0.68, whereas cDTW\textsuperscript{SS} had an average RI of 0.67. We note that the claim in \cite{KeoghSS} was based on a comparison on two datasets. Our experiments clearly indicate that it does not generalize towards all datasets.

Thus, contrary to earlier suggestions, our results indicate that constraints are better used to select and tune the algorithm (i.e.\, COBS) than to tune the similarity measure (i.e.\ cDTW\textsuperscript{SS}).

\subsubsection*{Runtime} 
COBRAS\textsuperscript{DTW}, cDTW\textsuperscript{SS} and COBS require the construction of the pairwise DTW distance matrix. This becomes infeasible for large datasets. For example, computing one distance matrix for the ECG5000 dataset took ca. 30h in our experiments, using an optimized C implementation of DTW. 

k-Shape and k-MS are much more scalable \cite{kshape}, as they do not require computing a similarity matrix. COBRAS\textsuperscript{k-Shape} inherits this scalability, as it uses k-Shape to refine super-instances. In our experiments, COBRAS\textsuperscript{k-Shape} was on average 28 times faster than COBRAS\textsuperscript{DTW}. 

\section{Case studies: CBF, TwoLeadECG and MoteStrain}
\label{sec:casestudies}

In this section, we investigate why COBRAS\textsuperscript{TS} significantly outperforms its competitors. Our main claim is that COBRAS\textsuperscript{TS} is able to deal with the inherent complexity of time series clustering by repeatedly refining super-instances. 

To support this claim, we inspect the clusterings that are generated for three UCR datasets in more detail. CBF and TwoLeadECG are examples for which COBRAS\textsuperscript{DTW} and COBRAS\textsuperscript{k-Shape} significantly outperform their competitors, whereas MoteStrain is one of the few datasets for which they are significantly outperformed by unsupervised k-Shape clustering.
These three datasets illustrate different reasons why time series clustering may be difficult. Clustering CBF is difficult because of the fact that one of the clusters comprises two separated subclusters; TwoLeadECG, because only limited \emph{subsequences} of the time series are relevant for the clustering at hand, and the remaining parts obfuscate the distance measurements; and MoteStrain, because of \emph{noise}. 

\subsection*{CBF}

\begin{figure}
\centering     
\includegraphics[width=1.0\textwidth]{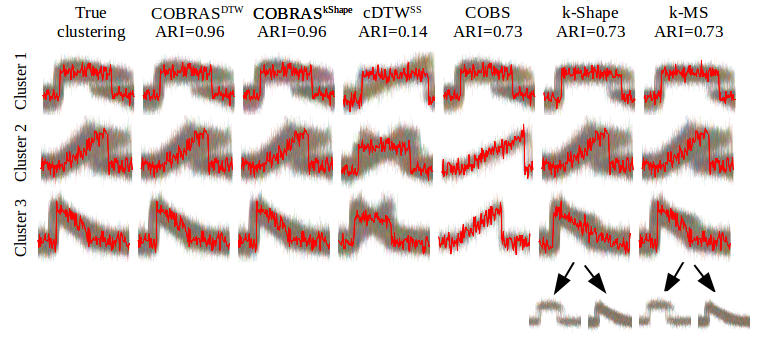}
\caption{The first column shows the true clustering of CBF. The remaining columns show the clusterings that are produced by all considered methods. For COBS, only the three largest of 15 clusters are shown. The prototypes are shown in red. For COBRAS\textsuperscript{DTW}, cDTW\textsuperscript{SS} and COBS the prototypes are selected as the medoids w.r.t.\ DTW distance. For COBRAS\textsuperscript{k-Shape}, k-Shape and k-MS the prototypes are the medoids w.r.t.\ the SBD distance.}\label{fig:cbf_comparison}
\end{figure}

The first column of Figure \ref{fig:cbf_comparison} shows the ``true'' clusters as they are indicated by the class labels. It is clear that the  classes correspond to three distinct patterns (horizontal, upward and downward). The next columns show the clusterings that are produced by each of the competitors. Semi-supervised approaches are given a budget of 50 queries. COBRAS\textsuperscript{DTW} and COBRAS\textsuperscript{k-Shape} are the only methods that provide a near perfect solution (ARI = 0.96). cDTW\textsuperscript{SS} mixes patterns of different types in each cluster. COBS find pure clusters, but too many: the plot only shows the largest three of 15 clusters for COBS. k-Shape and k-MS mix horizontal and downward patterns in their third cluster. To clarify this mixing of patterns, the figure shows the instances in the third k-Shape and k-MS clusters again, but separated according to their true class.

\begin{figure}
\centering     
\includegraphics[width=0.7\textwidth]{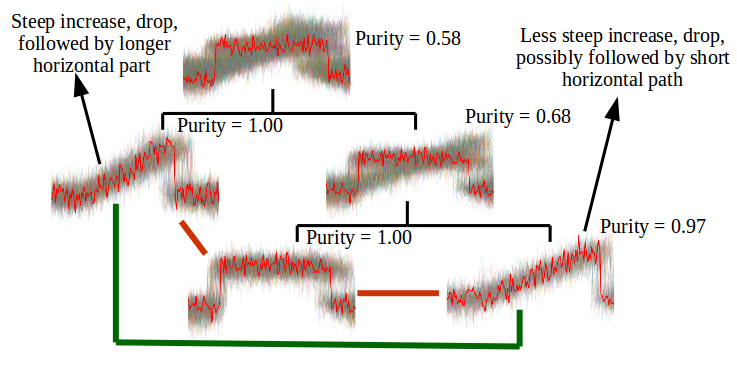}
\caption{A super-instance that is generated while clustering CBF, and its refinements. The green line indicates a must-link constraint, and illustrates that these two super-instances will be part of the same multi-modal cluster (corresponding to upward patterns). The red lines between super-instances indicate cannot-link constraints. The purity of a super-instance is computed as the ratio of the occurrence of the most frequent class in the super-instance, over the total number of elements in the super-instance.}\label{fig:superinstances}
\end{figure}

Figure \ref{fig:superinstances} illustrates how repeated refinement of super-instances helps COBRAS\textsuperscript{TS} deal with the complexities of clustering CBF. It shows a super-instance in the root, with its subsequent refinements attached as children. The super-instance in the root of Figure \ref{fig:superinstances} (which is itself a result of a previous super-instance split) contains time series showing horizontal and upward patterns. Clustering it into two new super-instances does not yield a clean separation of these two types: a perfectly pure cluster with upward patterns is created, but the other super-instance still mixes horizontal and upward patterns. This is not a problem for COBRAS\textsuperscript{TS}, as it simply refines the latter super-instance again. This time the remaining time series are split into nearly pure super-instances separating horizontal from upward patterns. Note that the two super-instances containing upward patterns correspond to two distinct subclusters: some upward patterns drop down very close to the end of the time series, whereas the drop in the other subcluster occurs much earlier. Typically, patterns in the latter subcluster increase with a steeper slope.


The clustering process just mentioned illustrates the point made earlier, in Section~\ref{sec:sepcomp}, about COBRAS's ability to construct clusters with separated components. It is clear that this ability is advantageous in the CBF dataset. Note that being able to deal with \emph{separated} components is key here; k-MS, which is able to find multi-modal clusters, but not clusters with modes that are separated by a mode from another cluster, produces a clustering that is far from perfect for CBF.


\subsection*{TwoLeadECG}

 \begin{figure*}
\centering     
\includegraphics[width=1.0\textwidth]{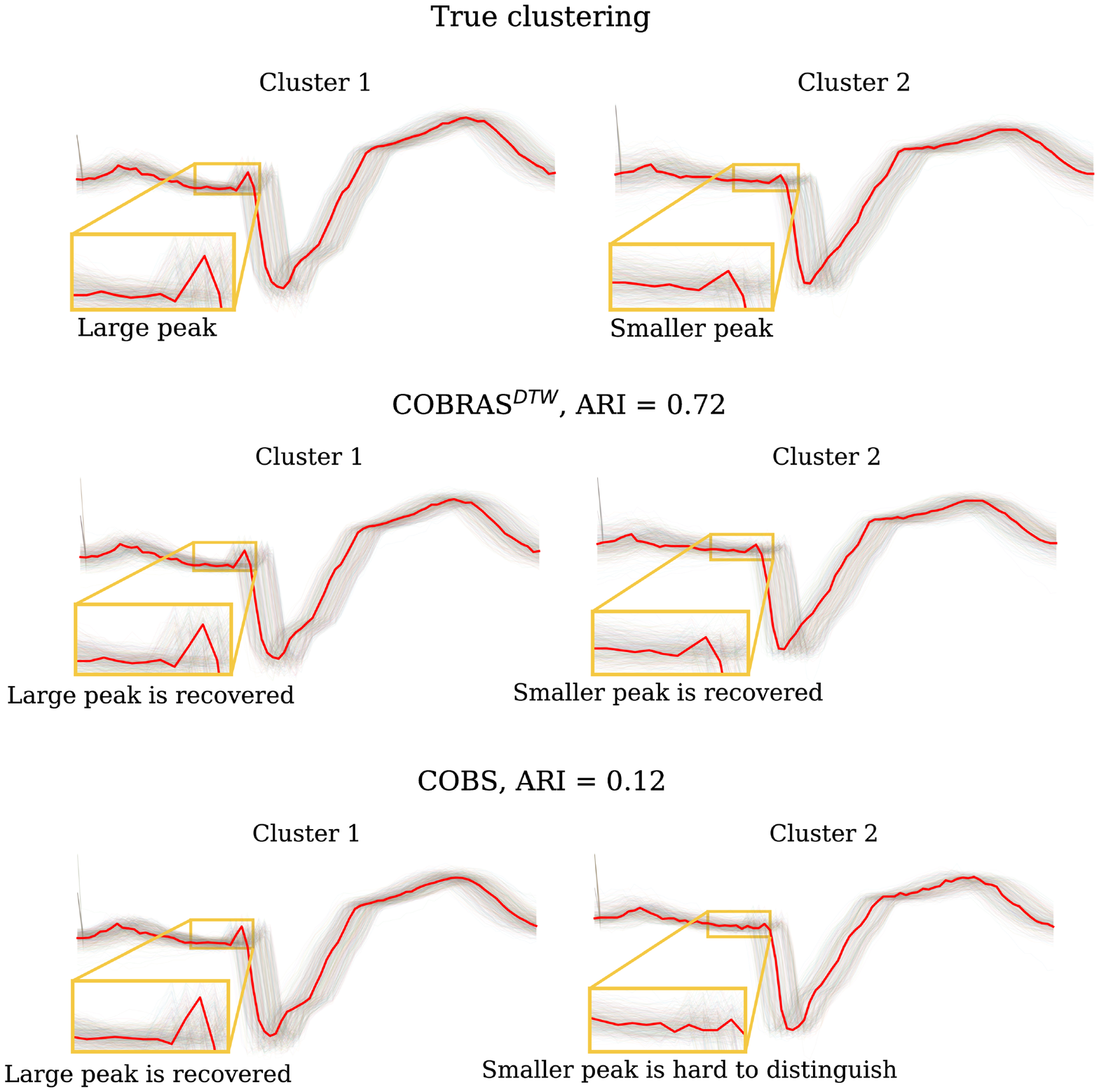}
\caption{The first column shows the ``true'' clustering of TwoLeadECG. The second column shows the clustering produced by COBRAS\textsuperscript{DTW}. The third column shows the clustering produced by COBS, which is the best competitor for this dataset. Prototypes are shown in red, and are the medoids w.r.t.\ the DTW distance.}\label{fig:twoleadecg_comparison}
\end{figure*}

The first column in Figure \ref{fig:twoleadecg_comparison} shows the ``true'' clusters for TwoLeadECG. Cluster 1 is defined by a large peak before the drop, and a slight bump in the upward curve after the drop. Instances in cluster 2 typically only show a small peak before the drop, and no bump in the upward curve after the drop. For the remainder of the discussion we focus on the peak as the defining pattern, simply because it is easier to see than the more subtle bump. 

The second column in Figure \ref{fig:twoleadecg_comparison} shows the clustering that is produced by COBRAS\textsuperscript{DTW}; the one produced by COBRAS\textsuperscript{k-Shape} is highly similar. They are the only methods able to recover these characteristic patterns. The last column in Figure \ref{fig:twoleadecg_comparison} shows the clustering that is produced by COBS, which is the best of the competitors. This clustering has an ARI of 0.12, which is not much better than random. From the zoomed insets in Figure \ref{fig:twoleadecg_comparison}, it is clear that this clustering does not recover the defining patterns: the small peak that is characteristic for cluster 2 is hard to distinguish.

This example illustrates that by using COBRAS\textsuperscript{TS} for semi-supervised clustering, a domain expert can discover more accurate explanatory patterns than with competing methods. None of the alternatives is able to recover the characteristic patterns in this case, potentially leaving the domain expert with an incorrect interpretation of the data. Obtaining these patterns comes with relatively little additional effort, as with a good visualizer answering 50 queries only takes a few minutes. This time would probably be insignificant compared to the time that was needed to collect the 1139 instances in the TwoLeadECG dataset. 

\begin{figure*}
\centering     
\subfigure[]{\label{fig:motestrain_si_1}\includegraphics[width=0.3\textwidth]{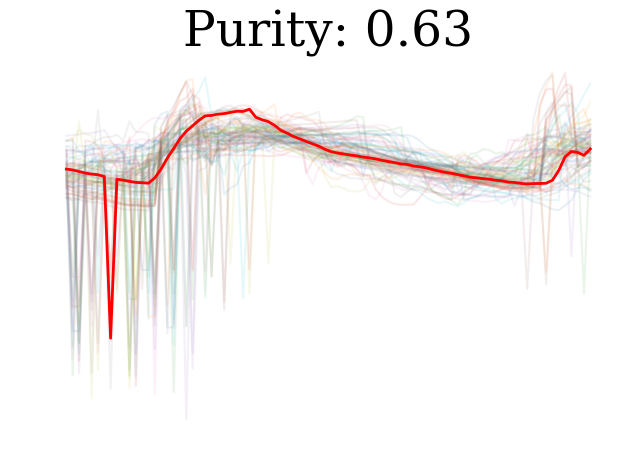}}
\subfigure[ ]{\label{fig:motestrain_si_2}\includegraphics[width=0.3\textwidth]{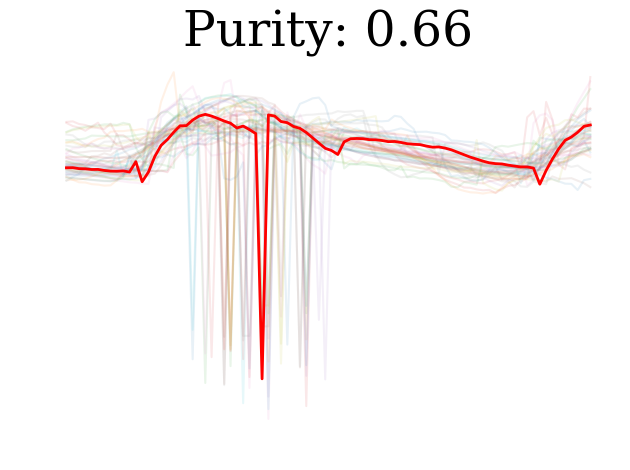}}
\caption{Two super-instances generated by COBRAS\textsuperscript{DTW}. The super-instances are based on the location of the noise.}\label{fig:motestrain_example}
\end{figure*}

\subsection*{MoteStrain}

In our third case study we discuss an example for which COBRAS\textsuperscript{TS} does not work well, as this provides insight into its limitations. We consider the MoteStrain dataset, for which the unsupervised methods perform best. k-MS attains an ARI of 0.62, and k-Shape of 0.61. COBRAS\textsuperscript{k-Shape} ranks third with an ARI of 0.51, and COBRAS\textsuperscript{DTW} fourth with an ARI of 0.48. These results are surprising, as the COBRAS algorithms have access to more information than the unsupervised k-Shape and k-MS. Figure \ref{fig:motestrain_example} gives a reason for this outcome; it shows  that COBRAS\textsuperscript{TS} creates super-instances that are based on the location of the noise. The poor performance of the COBRAS\textsuperscript{TS} variants can in this case be explained by their large variance. The process of super-instance refinement is much more flexible than the clustering procedure of k-Shape, which has a stronger bias. For most datasets, COBRAS\textsuperscript{TS}'s weaker bias led to performance improvements in our experiments, but in this case it has a detrimental effect due to the large magnitude of the noise. In practice, the issue could be alleviated here by simply applying a low-pass filter to remove noise prior to clustering.

\section{Conclusion}
\label{sec:conclusion}

Time series arise in virtually all disciplines. Consequently, there is substantial interest in methods that are able to obtain insights from them. One of the most prominent ways of doing this, is by using clustering. In this paper we have presented COBRAS\textsuperscript{TS}, an novel approach to time series clustering. COBRAS\textsuperscript{TS} is semi-supervised: it uses small amounts of supervision in the form of must-link and cannot-link constraints. This sets it apart from the large majority of existing methods, which are unsupervised. An extensive experimental evaluation shows that COBRAS\textsuperscript{TS} is able to effectively exploit this supervision; it outperforms unsupervised and semi-supervised competitors by a large margin. As our implementation is readily available, COBRAS\textsuperscript{TS} offers a valuable new tool for practitioners that are interested in analyzing time series data.  

Besides the contribution of the COBRAS\textsuperscript{TS} approach itself, we have also provided insight into why it works well. A key factor in its success is its ability to handle clusters with separated components.

\section*{Acknowledgements}
We thank Hoang Anh Dau for help with setting up the cDTW\textsuperscript{SS} experiments. Toon Van Craenendonck is supported by the Agency for Innovation by Science and Technology in Flanders (IWT). This research is supported by Research Fund KU Leuven (GOA/13/010), FWO (G079416N) and FWO-SBO (HYMOP-150033).

\bibliographystyle{abbrv}
\bibliography{sample}



\end{document}